# The Scorecard for Synthetic Medical Data Evaluation and Reporting

Ghada Zamzmi, Adarsh Subbaswamy, Elena Sizikova,
Edward Margerrison, Jana Delfino, Aldo Badano

Office of Science and Engineering Laboratories, Center for Devices and Radiological Health, Food and Drug Administration.

*Corresponding author(s). E-mail(s): alzamzmigaa@fda.hhs.gov;

**Abstract**

Although interest in synthetic medical data (SMD) for training and testing AI methods is growing, the absence of a standardized framework to evaluate its quality and applicability hinders its wider adoption. Here, we outline an evaluation framework designed to meet the unique requirements of medical applications, and introduce SMD Card, which can serve as comprehensive reports that accompany artificially generated datasets. This card provides a transparent and standardized framework for evaluating and reporting the quality of synthetic data, which can benefit SMD developers, users, and regulators, particularly for AI models using SMD in regulatory submissions.

## 1 Introduction

A key challenge for the safe and effective development of medical artificial intelligence (AI) devices is the limited availability of patient data [13], as data sharing is often restricted due to well-founded privacy concerns. Further, data collection is time-consuming, costly, and sometimes unfeasible for rare and underrepresented populations. Synthetic medical data (SMD)–artificial data partially or fully generated using computational techniques to mimic the properties and relationships seen in patient data [18]– holds promise for addressing these emerging challenges.

SMD has gained attention due to recent advances in generative deep learning techniques. Methods, such as Generative Adversarial Networks (GANs) and Denoising Diffusion Probabilistic Models, have the ability to approximate the complex distributions of medical data and create synthetic distributions that align with patient data.

These methods hold promise for producing large quantities of medical data at scale, which could supplement the scarce patient data currently available for medical AI development and evaluation. However, the usefulness of SMD hinges on its quality. If SMD is not curated carefully, it could result in poor outcomes of downstream tasks – a classic case of "garbage in, garbage out".

Therefore, a systematic approach is needed for evaluating and reporting the quality of SMD to enable its proper use in medical applications. Current methods [1] for assessing synthetic data are inadequate because they do not holistically evaluate the generated data based on medically relevant criteria. Evaluation strategies commonly used are based on metrics initially developed for computer vision tasks, focusing primarily on visual and statistical fidelity [1]. This focus on fidelity alone fails to consider the unique and complicated aspects of medical data. For example, Figure 1 displays synthetically generated medical images that perform well on known metrics such as *Fréchet Inception Distance (FID)* and statistical similarity measures. Nevertheless, these images fail to meet constraints (e.g., broken ligaments, or multiple nipples) and exhibit artifacts such as misplaced pacemakers. While these artifacts may pass undetected by existing evaluation metrics, they risk misleading AI models trained or tested on such data, potentially resulting in incorrect predictions.

The figure also highlights current issues with evaluating synthetic textual data generated by large language models (LLMs). In the medical query task, the LLM provided incorrect answers. As current evaluation metrics [2] (e.g., *BLEU* and *ROUGE*) are designed to assess the similarity between generated text and ground truth responses based on n-gram overlap, these metrics can successfully detect explicit mismatches, such as incorrect answers, by measuring how closely the generated text matches the reference text. However, these same metrics fail to account for hallucinated or fabricated information [2], as demonstrated in the summarization task. In this task, the LLM introduces a hallucinated symptom (leg swelling) that is not present in the original Electronic Health Record (EHR) note. This error exposes a fundamental limitation of BLEU and ROUGE as they focus on overlap and recall of n-grams without evaluating the factual accuracy or contextual relevance of the generated text. Even if the hallucinated symptom is medically irrelevant or nonsensical in the given context, these metrics fail to identify such discrepancies because they do not analyze content for clinical constraints. This limitation poses risks in clinical applications. For example, a clinician relying on the summary might assume the presence of a symptom explicitly denied by the patient, which can potentially lead to misdiagnosis or inappropriate treatment. These examples underscore the need for comprehensive evaluation frameworks that go beyond fidelity and statistical similarity to incorporate dimensions such as clinical constraints (e.g., clinical validity and anatomical accuracy).

In this paper, we present a framework (Figure 2) for evaluating and reporting SMD based on seven medically relevant criteria and introduce SMD Card. Such a card would accompany artificially generated medical dataset, and provide quantitative assessments of SMD quality across multiple criteria while documenting key information such as dataset description, intended use, limitations, recommendations, and usage disclaimers. We encourage data developers, providers, and users to adopt this practice to promote rigorous evaluation, enhance transparency, and build trust in synthetic datasets.

## 1.1 Approaches for Evaluating Synthetic Medical Data

We categorize approaches for evaluating SMD into three main types: intrinsic data quality evaluation (task-independent), task-dependent evaluation, and human-dependent evaluation (Table 1).

Intrinsic data quality evaluation focuses on assessing the intrinsic quality of the synthetic dataset itself, independent of any specific downstream task. This involves evaluating the data in the image, raw data, or feature space based on criteria such as congruence, coverage, and constraint adherence, as outlined in the evaluation framework. This approach allows for an early indication of potential quality issues that could later affect task-specific performance. For instance, if general data quality evaluation reveals low fidelity or inadequate coverage, this suggests that the dataset may lack critical patterns or diversity, which might result in poor generalization and subpar performance in downstream applications. The task-independent approach is advantageous for identifying data quality issues early, but it requires robust feature representation and quantification to be effective.

Task-dependent evaluation measures the quality of synthetic data based on its effectiveness for a specific task. This method typically involves training a model on synthetic data and then testing it on real data, or vice versa, to assess whether the synthetic data leads to better training outcomes or improved testing results. This approach provides detailed insights into the utility of the synthetic data and helps determine its effectiveness in supporting the intended purpose. However, its generalizability is limited because it evaluates the synthetic data's quality solely within the context of that specific task. Furthermore, this approach tends to identify issues late in the evaluation process, which can delay necessary improvements. Such late detection is costly as it requires revisiting and retraining models to enhance the datasets. This process can become cyclical if the root causes of poor task performance are not fully understood. Consequently, task-independent evaluation, which involves a deeper and comprehensive assessment of the data based on well-defined criteria such as correctness, coverage, and constraints, is needed for ensuring broader data quality.

Human-dependent visual inspection is a subjective evaluation method, where domain experts visually compare synthetic and real data to assess quality. This approach is valuable for initial quality checks and can capture subtle, qualitative aspects that automated metrics might overlook. However, it is inherently subjective, inconsistent, and not scalable, as it relies on individual interpretation and may miss minor statistical deviations.

Each evaluation approach—intrinsic SMD quality, task-specific, and human-dependent—has unique strengths and limitations. For optimal assessment, these approaches should be combined rather than used in isolation. Starting with intrinsic data quality evaluation provides a broad overview and can uncover early issues that may affect downstream tasks. Following this with task-specific evaluations allows us to measure SMD utility, and check if early quality issues, such as inadequate coverage, align with task-specific performance limitations. Human-dependent visual inspection can further validate SMD. This combination ensures a comprehensive evaluation, where synthetic datasets are systematically refined through quantitative metrics, task performance, and human expertise.

**Table 1** Summary of Evaluation Approaches for SMD

| Approach | Scope | Task-Specific? | Subjective? | Identify Early Issues? | Scalable? |
|---|---|---|---|---|---|
| **Intrinsic Quality (7Cs)** | Dataset / Sample | × | × | ✓ | ✓ |
| **Task-Specific** | Dataset | ✓ | × | × | ✓ |
| **Human-Dependent** | Sample | × | ✓ | × | × |

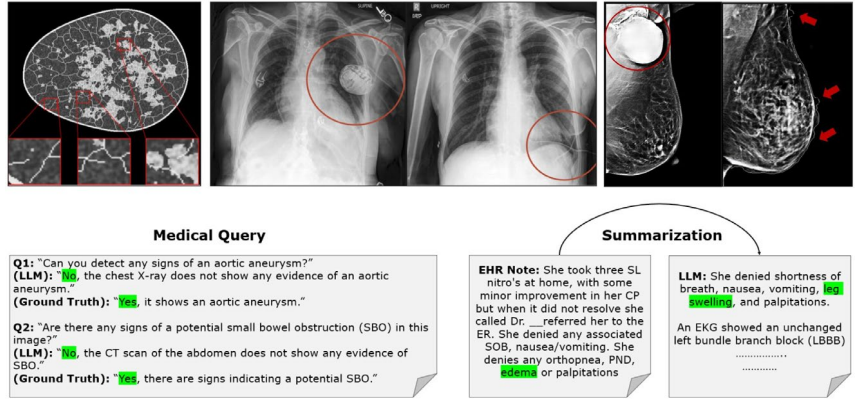

**Fig. 1** The first row highlights issues in synthetic images. The first image on the left (source: [6]) shows a synthetic image produced by a generative AI model; while the image achieves high visual fidelity, it contains structural inconsistencies, such as broken ligaments. The second and third images are synthetic chest X-rays (source: [7]), featuring misplaced medical devices like pacemakers and tubes located outside of anatomically plausible regions. The fourth and fifth images are digital mammograms that score highly on fidelity and statistical metrics but display clinically implausible artifacts, such as a bright circular region and the erroneous presence of multiple nipple-like structures. The second row focuses on issues in textual outputs from LLMs. In the medical query task, the LLM provides incorrect responses to critical clinical questions (source: [3]), which can often be detected through direct comparison with ground truth. However, in the summarization task (source: [19]), the LLM generates a symptom (leg swelling) not present in the original EHR note, which is harder to detect using similarity- or overlap-based metrics.

## 2 The "7 Cs" of SMD Evaluation

*"When a measure becomes a target, it ceases to be a good measure."*

– Goodhart's Law

In the context of SMD, this law underscores the risks of focusing exclusively on a single measure, such as alignment between synthetic and patient data. While such alignment is important, over-optimizing for it can result in neglecting other criteria. To address this, we propose evaluating SMD using a set of comprehensive criteria and measures. To address this, we propose evaluating SMD using a set of comprehensive criteria and measures. This section presents the "7 Cs": **C**orrectness, **C**overage, **C**onstraint, **C**ompleteness, **C**ompliance, **C**omprehension, and **C**onsistency.

While we recognize the value of qualitative approach for assessing SMD, we believe there is a need to establish quantitative approach to drive consistency and standard- ization. We also note that the aim of such evaluation is not to identify the 'best' synthetic dataset. Rather, the aim is to establish a transparent and comprehensive framework for assessing datasets across multiple clinically relevant criteria. Figure 2 illustrates the proposed criteria and evaluation card, with additional details about the Card provided in Table 2.

## Congruence

Congruence assesses the degree to which the distribution of synthetic data aligns with the distribution of patient data. High Congruence indicates that synthetic sam- ples closely match real ones in statistical properties and perceptual quality. In the literature, this concept is related to realism, which emphasizes perceptual quality, and fidelity, which focuses on the statistical alignment of generated and real data distributions [5, 12, 1].

Although Congruence alone is not sufficient to assess SMD, ensuring alignment between synthetic and real data is a necessary first step before evaluating other aspects. Congruence can be evaluated using several metrics tailored to different data types [1, 5]. For image data, metrics such as *Cosine Similarity* and *FID* assess alignment in the feature space and compare the distributions of synthetic and real images, respectively. For textual data, *Cosine Similarity* measures alignment between text embeddings, while metrics like the *BLEU* score evaluate similarity based on n-gram overlap between synthetic and reference texts [2]. These metrics are summarized under the Congruence dimension in Table 3.

## Coverage

Coverage evaluates the extent to which SMD captures the diversity and novelty inher- ent in patient data while aligning with true distributions. Diversity ensures that SMD represents the variability and breadth of patterns, features, and modes present in real data. Novelty evaluates whether the dataset contains unique samples that introduce new and valid variations while remaining aligned with real-world plausibility.

Coverage evaluates the diversity and novelty of synthetic data across various data types [1, 5]. For image data, metrics such as *Convex Hull Volume* and *Clustering-Based* metrics can quantify diversity by examining the spread and distribution of data points within the feature space. For textual data, *Recall* can assess Coverage by measuring the proportion of distinct terms or patterns in synthetic notes compared to real notes. For numerical data, metrics like *Variance* and *Entropy* evaluate the range and dispersion of synthetic values, ensuring they span the real-world distribution while introducing novel but medically valid combinations. Other metrics can be found in Table 3.

## Constraint

We define Constraint as the extent to which SMD respects known constraints, which may include anatomical, geometric, or clinical constraints.

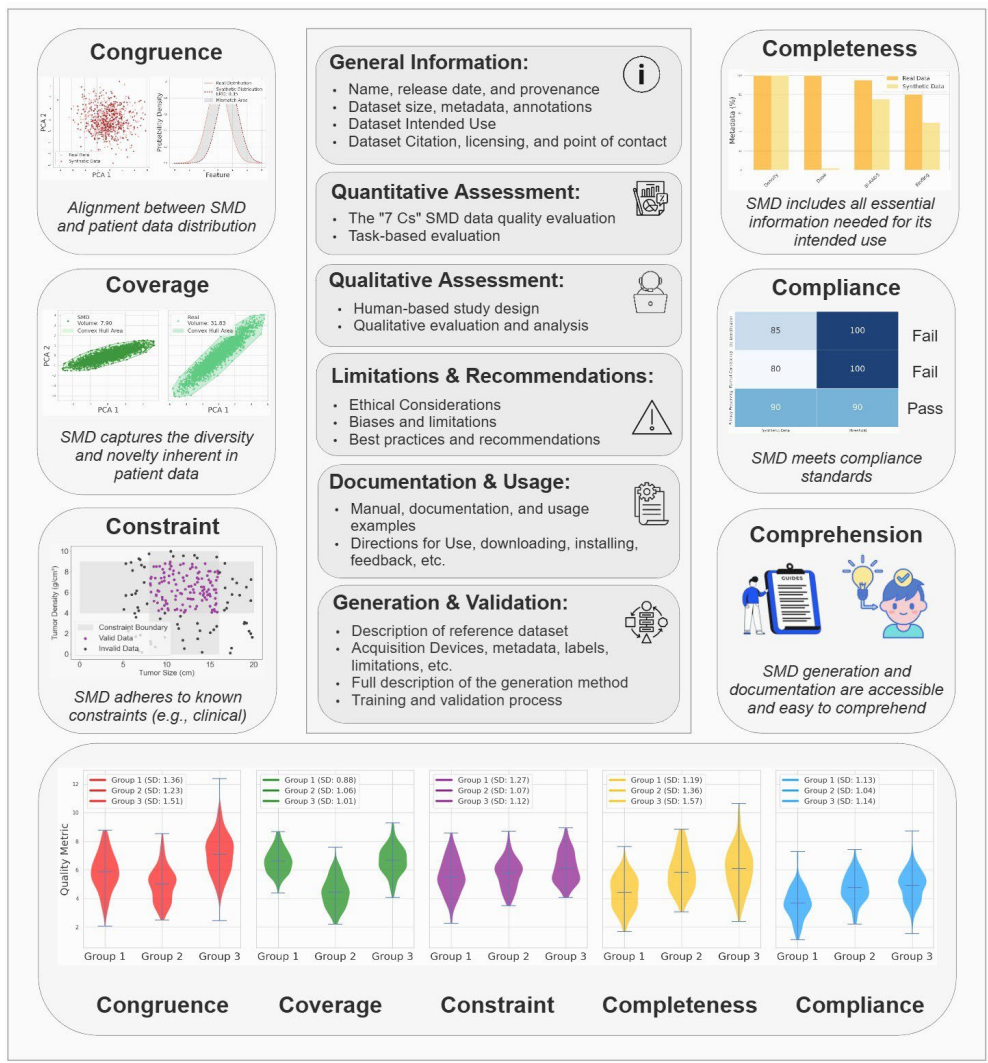

**Fig. 2** Framework for evaluating and reporting synthetic medical data based on seven criteria; each criterion is graphically depicted for intuitive visualization. Congruence measures the alignment between the distributions of synthetic and real data. Coverage highlights the diversity of synthetic data, demonstrated using convex hull volume, where synthetic data shows a smaller spread compared to patient data. Constraint evaluates adherence to clinical boundaries (e.g., tumor size and density) by identifying valid versus invalid data points that reside outside a defined boundary. Completeness assesses whether SMDs include all necessary information for their intended use. Compliance ensures adherence to standards such as privacy preservation, file format consistency, and data auditing. Comprehension evaluates the accessibility and clarity of SMD generation and documentation. The bottom panel illustrates the consistency of SMDs across subgroups, which can include demographic or disease-specific groups. Essential descriptive information about the synthetic data is presented in the middle panel.

**Table 2** Synthetic Medical Data (SMD) Card Template. Section 2 (Data Quality Evaluation) presents the quantitative results based on the "7 Cs" introduced in this work.

| 1. Synthetic Data General Information | |
|---|---|
| Name | [Dataset name] |
| Release Date | [Release date] |
| Version History | [Version numbers and updates] |
| Dataset Size | [Size, including number of samples, cases, and patients] |
| Dataset Modality | [Data modality, e.g., imaging (CT, MRI, X-ray), textual] |
| Dataset Provenance | [Origin and generation method of the dataset] |
| Dataset Intended Use | [Primary and secondary applications and use cases] |
| Dataset Labels | [Details of labels including types (e.g., binary, categorical, segmentation)] |
| Attribution and Licensing | [Citation information and license details] |
| Point of Contact | [Contact details for inquiries] |
| **2. Data Quality Evaluation ("7 Cs" Quantitative Results)** | |
| Congruence | [Evaluation of alignment between synthetic and real data] |
| Coverage | [Evaluation of feature and population diversity and representation] |
| Constraint | [Evaluation of adherence to known constraints (e.g., clinical or technical)] |
| Completeness | [Evaluation of missing or incomplete information in the synthetic data] |
| Compliance | [Evaluation of adherence to ethical, clinical, or regulatory standards] |
| Comprehension | [Evaluation of interpretability or transparency, if applicable] |
| Consistency | [Evaluation of data quality across batches, subgroups, or over time] |
| **3. Task-based Evaluation (Quantitative Results)** | |
| Task Performance | [Evaluation of the synthetic dataset's effectiveness for a specific task] |
| Task-Specific Metrics | [Metrics such as sensitivity, or mean IoU along with acceptance thresholds] |
| **4. Human-based Evaluation (Qualitative Results)** | |
| Human Study Design | [Summary of the reader study design, including evaluator expertise] |
| Reader Study Results | [Results of human evaluation (e.g., realism, and clinical relevance)] |
| Observations & Failure Cases | [Insights from qualitative evaluation, highlighting strengths, weaknesses, or failure cases] |
| **5. Ethical, Legal, and Practical Considerations** | |
| Privacy & Anonymization | [Details on anonymization, privacy, and compliance with standards] |
| Biases | [Known biases, their potential impacts on dataset use, and applied or suggested mitigation strategies] |
| Limitations | [Other technical, clinical, or regulatory limitations of the dataset] |
| Recommendations | [Recommendations and best practices] |
| **6. Synthetic Dataset Usage** | |
| Repository Access | [Direct link to the dataset repository, DOI, and any access requirements] |
| Preprocessing Requirements | [Details of data format and preprocessing steps required for use] |
| User Documentation | [Links to README files, user guides, tutorials, or other supporting materials for dataset usage] |
| Intended Audience | [Description of target users (e.g., researchers, clinicians, developers)] |
| **7. Synthetic Dataset Training & Validation Process** | |
| Generation Method | [Description of the synthetic data generation pipeline] |
| Training & Validation Process | [Details of the training and validation process] |
| **8. Reference Dataset General Information** | |
| Purpose | [Description of the reference dataset's purpose and primary use cases] |
| Origin & Source | [Origin of the reference dataset, sites, and dates from which data was collected] |
| Dataset Size | [Details on the size including number of samples, cases, or patients] |
| Clinical Population | [Characteristics of the clinical population, such as demographics, disease prevalence, and inclusion/exclusion criteria] |
| Acquisition Devices | [Details of the devices and parameters used for data acquisition] |
| Reference Standard | [Details of the reference standard, including annotation methods and qualifications of annotators] |
| Ground Truth Labels | [Details of ground truth annotations available] |
| Metadata | [Available metadata such as age, gender, breast density, etc.] |
| Preprocessing | [Preprocessing steps applied to the dataset] |
| Known Limitations | [Known limitations of the reference dataset such as biases or technical constraints] |

Constraint violations can occur in both imaging and non-imaging synthetic data. For instance, in the summarization task (Figure 1), a hallucinated symptom is intro- duced that is absent from the original EHR note. In digital mammography (Figure 1), maintaining anatomical accuracy requires adherence to constraints such as correct breast shape and the appropriate number of nipples. Additionally, enforcing con- straints such as size and density in synthetic tumor images or ensuring valid lab value ranges in generated medical reports is important for making SMD suitable for training diagnostic tools.

Constraint adherence can be quantified using various metrics. The *Constraint Violation Rate* (see Table 3) measures the frequency at which generated data devi- ates from predefined clinical constraints or fails to align with relevant symptoms for specific medical conditions. Metrics such as *Nearest Invalid Datapoint* and *Distance to Constraint Boundary* evaluate the proximity of synthetic data points to clini- cally established boundaries, ensuring that generated data remains within acceptable ranges. In some cases, these constraint boundaries can be directly established from patient data to reflect real-world distributions. Other metrics, such as those evaluat- ing structural and anatomical plausibility (e.g., maintaining realistic tissue density or geometric accuracy), are detailed in [10, 16].

## Completeness

Completeness is defined as the extent to which generated data contains all necessary details relevant to the task. In medical imaging, Completeness can refer to whether essential information, such as metadata, annotations, and clinical details, is included when required for the application. For textual data, Completeness ensures that the generated output provides complete information and findings without omitting key elements essential for the clinical task. In tabular data, Completeness involves includ- ing all relevant features and records needed for the clinical task; missing data or incomplete records would lead to lower Completeness.

Given a reference standard for comparison, several metrics can quantify Complete- ness. For example, the *Proportion of Required Fields* (Table 3) measures the fraction of essential fields present in the synthetic data relative to the reference dataset. *Miss- ing Data Percentage* quantifies the proportion of missing values compared to the total number of expected values. Additionally, scaling-based metrics provide a measure of Completeness by rating how well the generated data captures essential details on a defined scale (e.g., 1 to 10). These metrics quantify the extent to which SMD pre- serves the necessary details found in patient data, which help in identifying missing information that could compromise the usability of the SMD.

## Compliance

We define Compliance as the adherence of SMD to established format guidelines (e.g., DICOM for medical imaging), privacy standards, and relevant local and international regulations. Compliance is particularly important in data-driven methods that might suffer from leakage of protected patient information. For example, GANs and LLMs are designed to generate SMD that closely approximates real-world datasets. However, when these models are trained on extensive datasets without adequate safeguards, they may inadvertently reproduce outputs containing sensitive patient information [8]. A significant privacy concern arises when the generated data is overly similar to specific real-world entries, potentially replicating identifiable details from the training data and leading to privacy violations.

Metrics to measure privacy compliance include *Differential Privacy* metrics, which evaluate the level of privacy protection in the synthetic data, and *Re-identification Risk* metrics, which assess the likelihood of identifying individuals in the synthetic data. Also, audit trails and compliance reports can be used to continuously monitor and document the adherence of SMD generation to standards. See Table 3 for examples of Compliance metrics, and refer to [8] for further details on privacy-preserving methods in synthetic data generation.

We recognize that privacy and regulatory requirements vary across regions and institutions, and the Compliance criterion is broad enough to accommodate diverse contexts. For example, in the U.S., Compliance would emphasize HIPAA (Health Insurance Portability and Accountability Act) standards for privacy, while in Europe, the focus would shift to GDPR (General Data Protection Regulation. Despite these regional differences, Compliance can be consistently evaluated using the metrics in Table 3 with region-specific thresholds. For instance, some regions may require stricter thresholds (e.g., lower $\varepsilon$ or higher $k$) to minimize re-identification risks, while other regions allow more flexible thresholds. Regardless of the region, the same Compli- ance metrics can be applied, with thresholds adjusted to meet specific regulatory requirements.

## Comprehension

Comprehension refers to how easily users can understand the process of generating SMD. This criterion evaluates the transparency and clarity of the data generation process as well as the quality of the accompanying documentation. One metric to evaluate the quality of documentation is the *Documentation Clarity Score* (Table 3). This metric quantifies how clearly the generation process is documented by using scale-based methods.

Knowledge-based approaches for SMD generation typically achieve higher Comprehension because parameters such as tumor sizes, noise ranges, or dose levels are explicitly defined and well-documented. These methods often include detailed annotations and examples that directly link the synthetic data to underlying clinical rules or constraints, making them inherently more transparent and easier to understand and document. In contrast, generative AI methods, which rely on implicit processes, often score lower on Comprehension due to challenges in interpreting hidden parameters and the complex, opaque nature of their algorithms.

## Consistency

Consistency refers to the stability and uniformity of SMD quality metrics across different subgroups or over time. This criterion evaluates whether SMD consistently meets quality criteria (e.g., Congruence, Coverage, Constraint, Completeness) across diverse patient demographics, disease classes, or other subgroups. Ensuring uniform quality across these groups prevents disparities that could negatively impact downstream applications. Consistency also applies over time, ensuring that SMD quality remains stable as real-world data landscapes evolve. This is particularly important for applications such as longitudinal studies and disease progression modeling, where temporal stability in synthetic data is needed for reliable predictions.

To assess Consistency, metrics such as *Variance*, *Maximum-Minimum Difference*, and *Analysis of Variance* (ANOVA) (Table 3) can be employed. While *Variance* and *Maximum-Minimum Difference* quantify average and extreme variations across subgroups or time points, ANOVA is useful for statistically assessing whether the variability observed between different subgroups or conditions is significant.

## 3 Synthetic Data Card

Inspired by the *Model Card* [11] and *Healthsheet* [15], we propose the *Synthetic Data Card*. While the SMD Card shares similarities with the *Model Card* and *Healthsheet* in promoting documentation and standardization, it is uniquely tailored for synthetic datasets, emphasizing quality evaluation through the 7Cs framework alongside comprehensive reporting of dataset characteristics and limitations. This card serves as both a quality assurance tool and a resource for stakeholders—including developers, regulators, and researchers—to understand the characteristics, quality and limitations of SMD.

As shown in Table 2, this Card can include the following sections.

1. **General Information**: This section provides basic details about the dataset, such as modality, size, labels, licensing, and point of contact for inquiries.
2. **Data Quality Evaluation (Quantitative)**: The core of the SMD Card, referred to as the *Scorecard*, evaluates the dataset against the "7 Cs" criteria. This section of the Card presents detailed results on data quality and provides concrete conclusion for each dimension.
3. **Task-based Evaluation (Quantitative)**: This section provides quantification of the dataset's utility in a specific task using task-specific metrics.
4. **Human-based Evaluation (Qualitative)**: This section provides results from qualitative evaluations. It includes description of study design, results of the reader study, and a summary of identified weaknesses or failure cases.
5. **Ethical Considerations, Limitations, and Recommendations**: This section highlights ethical concerns including known biases and limitations. Recommendations for best practices are provided.
6. **Usage**: This section offers guidance on utilizing the dataset. It includes links to repositories or DOIs, details on preprocessing, documentation (e.g., README or user guides).
7. **Synthetic Dataset Training and Validation Process**: This section documents the processes used for generating and validating the synthetic dataset.
8. **Reference Dataset Information**: This section outlines key details of the real dataset used for comparison, including its size, population characteristics, and known limitations. Such information is particularly important for generative methods, as it forms the foundation for assessing the quality and relevance of the synthetic dataset.

This Card, combined with the "7 Cs" framework, can benefit various stakeholders. For example, regulatory agencies can leverage this transparent and standardized approach to understand the quality of SMD used for training or testing AI models submitted for regulatory approval. Data developers can use the card to provide comprehensive documentation and evaluation for their datasets. Researchers can use the Card to decide whether the SMD is suitable for specific applications or tasks based on its characteristics, including alignment with patient data, diversity, adherence to application-specific constraints, consistency across subgroups, compliance with relevant standards, and completeness to ensure no task-related information is missing.

To facilitate the creation of the Card, we developed a user-friendly webpage that simplifies the process of completing the descriptive sections. The webpage allows users to retrieve descriptive information from a file and summarize it into a Card template.

Additionally, we created a Python library, SMD-ScoreCard, to compute the quantitative aspects of the Card, focusing on the metrics associated with the "7 Cs" framework. We encourage researchers and developers to create their own versions of the Card or extend our implementation to address specific needs.

## 4 Discussion

The proposed SMD Card, along with its "7 Cs" quantitative framework offers a multimodal approach for evaluating the quality of SMD. Beyond establishing distribution alignment between synthetic and patient data (Congruence), the framework evalu- ates SMD based on other criteria. Coverage measures the diversity and variability of synthetic data to ensure it reflects real-world heterogeneity and trends. Constraint assesses adherence to known constraints (clinical, technical, geometrical). These three criteria address alignment, representation, diversity, and adherence to constraints. The framework also includes Completeness to assess that the dataset includes all the necessary information required for its intended application. Compliance verifies adherence to privacy regulations and standards. Consistency ensures that data quality remains uniform across different subgroups or over time. Lastly, Comprehension ensures that the dataset is accompanied by clear documentation.

This framework can be extended to different contexts and data types beyond medical imaging. For example, for clinical notes, Congruence can assess alignment using semantic features (e.g., token embeddings) and similarity metrics, while for tabular data, it compares statistical distributions of key attributes. Coverage can evaluate diversity in clinical notes via token variability, and in tabular data through statistical spread tests.

Similarly, Constraint can be applied to other data types. For example, in medical imaging, it ensures anatomical plausibility, such as verifying that breast mass sizes and shapes align with clinically plausible ranges using features like shape descriptors or spatial consistency checks. For clinical notes and tabular data, constraints can be derived from diagnostic guidelines or known physiological ranges. For instance, an important constraint in clinical notes is ensuring consistency in symptom-condition relationships, which can be achieved through graph modeling of extracted tokens to verify that symptom descriptions align with the documented condition. In numerical data, constraints can assess whether values fall within known physiological ranges; for example, hemoglobin levels should adhere to normal or condition-specific thresholds (e.g., ranges for anemia versus normal levels). By defining valid constraint boundaries—whether anatomical, clinical, or physiological—synthetic data can be rigorously validated.

Other criteria, such as Completeness and Consistency, can also be generalized. Consistency is inherently adaptable, as it focuses on assessing the uniformity of qual- ity criteria (e.g., Coverage, Constraints) across subgroups or over time. Completeness ensures the inclusion of all essential elements across data types: for imaging data, it can verify the presence of annotations and metadata; for textual data, it can ensure generated outputs include all critical information; and for numerical data, it validates the presence of all necessary variables and records. Finally, the Compliance criterion can emphasize core principles (e.g., data security and privacy) while adapting thresh- olds for standardized metrics to regional regulations. For example, privacy and data anonymization methods may use consistent metrics globally, but thresholds or specific requirements can vary by region. This approach ensures the metrics remain applicable while accommodating regional requirements effectively.

The quantitative assessment of certain criteria using specific metrics (see Table 3) requires transforming data into a feature embedding space. For images, common feature representations include radiomic features, deep features, topological features, and shape-based properties. For textual data, token embeddings are often used, while for numerical data, statistical summaries (e.g., mean, variance, clustering features) can serve as key features. Since high-dimensional feature spaces may contain redundant or irrelevant information, dimensionality reduction techniques (e.g., principal component analysis) or feature summarization methods are often applied to condense the data into compact embeddings. Such careful feature representation is important for accurate quantitative evaluation. Further, to achieve a deeper evaluation of data quality, we recommend assessing the quality of SMD at both global and local levels. Global assessments ensure that the synthetic data maintains high overall quality, capturing aspects such as global distribution alignment, diversity, and adherence to constraints. Local assessments, on the other hand, focus on specific regions of interest (e.g., local- ized lesion areas). For instance, high global Congruence indicates overall alignment between synthetic and patient data distributions (e.g., images exhibit similar texture, contrast, statistical properties), while high local Congruence ensures that the charac- teristics of lesion areas in synthetic data closely align with those in patient data. This hierarchical approach balances broad trends and nuanced details.

To obtain concrete conclusions for each criterion, metrics can be aggregated into a single score. Aggregate scoring is widely utilized across various domains (e.g., credit scoring) and can be effectively applied here. For instance, metrics within Coverage can be combined using methods such as normalized or weighted averages, geometric means, or other approaches. After calculating the aggregated score, thresholding can be applied. These thresholds can be informed by clinical knowledge or derived from a reference dataset. For example, if a reference patient dataset demonstrates a Coverage range between 80 and 85, the following thresholds could classify Coverage as: good if the score ≥ 80, moderate if 70 ≤ score < 80, and low if the score < 70. We note that the thresholds would vary depending on the clinical application or task. We also recommend reporting both the aggregated score and the individual metrics for each criterion

Ensuring the quality of SMD is necessary for its successful use, not only during AI testing but also at the training stage. Synthetic data has the potential to enhance performance by enlarging training sets, but poorly constructed or unrealistic data can introduce biases, confuse the model, and lead to the learning of irrelevant features, ultimately impairing the model's ability to generalize to real-world scenarios. This issue was exemplified in the failure of IBM's Watson for Oncology. This high-profile collapse was attributed to the reliance on synthetic data that failed to capture the complexities and variabilities of patient data. Similarly, a recent study [17] revealed that LLMs tend to degrade when trained on recursively generated data, further emphasizing the need to scrutinize the quality of synthetic data used in training and testing. Another study [4] discussed how synthetic data can compromise utility and affect downstream classification performance. To mitigate these issues, proactive quality evaluation is important for identifying and addressing potential limitations in the data early in the process.

With the growing reliance on synthetic data, the need for standardized and proactive approaches to evaluate its quality becomes increasingly important. These approaches enable researchers and practitioners to verify that synthetic data meets quality standards and aligns with its intended use. Additionally, as the demand for transparency and standardization continues to rise, we anticipate that journals, fund- ing agencies, and regulatory bodies will increasingly require comprehensive evaluations and reporting of SMD in the future.

**Table 3** Examples of metrics to evaluate SMD quality across seven dimensions. Implementations and further details about these metrics are available in our SMD ScoreCard library. The table includes the following columns: **Space**, indicating the domain where the metric operates (e.g., embedding, metadata fields); **Binary?**, specifying whether the metric can be applied to synthetic data alone (unary) or requires comparison with patient data (binary); **Direction**, describing the optimization goal (e.g., maximize, minimize, or statistical significance); and **Image-Metric?**, which indicates if the metric is restricted to image or signal data, or applicable to other types, such as clinical notes or numerical data. Note that **Direction** of Coverage metrics should be maximized within a specific range to ensure SMD is diverse while remaining valid and aligned with patient data. For further details about these metrics and how they are measured, refer to [1, 14, 9].

| Metric Example | Space | Binary? | Direction | Image Metric? |
|---|---|---|---|---|
| **Congruence** | | | | |
| Cosine Similarity | Embedding | Binary | Maximize | No |
| Earth Mover's Distance | Embedding | Binary | Minimize | No |
| Jensen-Shannon Divergence | Embedding | Binary | Minimize | No |
| Peak Signal-to-Noise Ratio | Image | Binary | Maximize | Yes |
| Structural Similarity Index | Image | Binary | Maximize | Yes |
| Fréchet Inception Distance | Embedding | Binary | Minimize | Yes |
| Distance to Centroid | Embedding | Binary | Minimize | No |
| Precision | Embedding | Binary | Maximize | No |
| **Coverage** | | | | |
| Inception Score | Image | Unary | Maximize | Yes |
| Recall | Embedding | Binary | Maximize | No |
| Coverage | Embedding | Binary | Maximize | No |
| Distance to Centroid | Embedding | Binary | Maximize | No |
| Convex Hull Volume | Embedding | Unary | Maximize | No |
| Determinantal Point Processes Score | Embedding | Unary | Maximize | No |
| Vendi Score | Embedding | Unary | Maximize | No |
| Variance | Embedding | Unary | Maximize | No |
| Entropy | Embedding | Unary | Maximize | No |
| Rarity Score | Embedding | Binary | Minimize | No |
| Clustering-Based Metrics | Embedding | Unary | Maximize | No |
| **Constraint** | | | | |
| Nearest Invalid Datapoint | Embedding | Binary | Minimize | No |
| Distance to Constraint Boundary | Embedding | Binary | Minimize | No |
| Constraint Violation Rate | Embedding | Binary | Minimize | No |
| **Completeness** | | | | |
| Proportion of Required Fields | Metadata | Binary | Maximize | No |
| Missing Data Percentage | Metadata | Binary | Minimize | No |
| **Compliance** | | | | |
| Differential Privacy Score | Data Attribute | Unary | Minimize | No |
| K-Anonymity Level | Data Attribute | Unary | Maximize | No |
| L-Diversity Score | Data Attribute | Unary | Maximize | No |
| T-Closeness Level | Data Attribute | Unary | Maximize | No |
| **Comprehension** | | | | |
| Documentation Clarity Score | Documentation | Unary | Maximize | No |
| **Consistency** | | | | |
| Variance | Quality Metrics | Unary | Minimize | No |
| Maximum-Minimum Difference | Quality Metrics | Unary | Minimize | No |
| Analysis of Variance | Quality Metrics | Unary | Stat. Sig. | No |

# References

[1] Ali Borji. "Pros and cons of GAN evaluation measures: New developments". In: *Computer Vision and Image Understanding* 215 (2022), p. 103329.

[2] Yupeng Chang et al. "A survey on evaluation of large language models". In: *ACM Transactions on Intelligent Systems and Technology* 15.3 (2024), pp. 1–45.

[3] Jiawei Chen et al. "Detecting and Evaluating Medical Hallucinations in Large Vision Language Models". In: *arXiv preprint arXiv:2406.10185* (2024).

[4] Victoria Cheng et al. "Can you fake it until you make it? impacts of differentially private synthetic data on downstream classification fairness". In: *Proceedings of the 2021 ACM Conference on Fairness, Accountability, and Transparency*. 2021, pp. 149–160.

[5] Fida K Dankar, Mahmoud K Ibrahim, and Leila Ismail. "A multi-dimensional evaluation of synthetic data generators". In: *IEEE Access* 10 (2022), pp. 11147–11158.

[6] Rucha Deshpande et al. "Assessing the capacity of a denoising diffusion probabilistic model to reproduce spatial context". In: *arXiv preprint arXiv:2309.10817* (2023).

[7] August DuMont Schütte et al. "Overcoming barriers to data sharing with medical image generation: a comprehensive evaluation". In: *NPJ digital medicine* 4.1 (2021), p. 141.

[8] Diogo André França Fernandes. "Synthetic data and re-identification risks". PhD thesis. Universidade do Porto (Portugal), 2022.

[9] Dan Friedman and Adji Bousso Dieng. "The vendi score: A diversity evaluation metric for machine learning". In: *arXiv preprint arXiv:2210.02410* (2022).

[10] Giorgio Giannone et al. "Learning from Invalid Data: On Constraint Satisfaction in Generative Models". In: *arXiv preprint arXiv:2306.15166* (2023).

[11] Margaret Mitchell et al. "Model cards for model reporting". In: *Proceedings of the conference on fairness, accountability, and transparency*. 2019, pp. 220–229.

[12] Muhammad Ferjad Naeem et al. "Reliable fidelity and diversity metrics for generative models". In: *International Conference on Machine Learning*. PMLR. 2020, pp. 7176–7185.

[13] Nariman Noorbakhsh-Sabet et al. "Artificial intelligence transforms the future of health care". In: *The American journal of medicine* 132.7 (2019), pp. 795–801.

[14] Lyle Regenwetter et al. "Beyond statistical similarity: Rethinking metrics for deep generative models in engineering design". In: *arXiv preprint arXiv:2302.02913* (2023).

[15] Negar Rostamzadeh et al. "Healthsheet: development of a transparency artifact for health datasets". In: *Proceedings of the 2022 ACM Conference on Fairness, Accountability, and Transparency*. 2022, pp. 1943–1961.

[16] Daniel Schaudt et al. "A Critical Assessment of Generative Models for Synthetic Data Augmentation on Limited Pneumonia X-ray Data". In: *Bioengineering* 10.12 (2023), p. 1421.

[17] Ilia Shumailov et al. "AI models collapse when trained on recursively generated data". In: *Nature* 631.8022 (2024), pp. 755–759.

[18] Elena Sizikova et al. "Synthetic data in radiological imaging: Current state and future outlook". In: *BJR— Artificial Intelligence* (2024), ubae007.

[19] Prathiksha Rumale Vishwanath et al. "Faithfulness Hallucination Detection in Healthcare AI". In: *Artificial Intelligence and Data Science for Healthcare: Bridging Data-Centric AI and People-Centric Healthcare*.